\pdfoutput=1

\documentclass[11pt]{article}

\usepackage{acl}

\usepackage{times}
\usepackage{latexsym}

\usepackage[T1]{fontenc}

\usepackage[utf8]{inputenc}

\usepackage{microtype}

\usepackage{inconsolata}

\usepackage{url}            
\usepackage{booktabs}       
\usepackage{amsfonts}       
\usepackage{nicefrac}       
\usepackage{xcolor}
\usepackage{multirow}
\usepackage[utf8]{inputenc} 
\usepackage[T1]{fontenc}    
\usepackage{hyperref}       
\usepackage{url}            
\usepackage{booktabs}       
\usepackage{amsfonts}       
\usepackage{nicefrac}       
\usepackage{xcolor}
\usepackage{microtype}      
\usepackage{amsmath}
\usepackage{xcolor}         
\usepackage{graphicx}
\usepackage{wrapfig}
\usepackage{algorithmic}

\usepackage{enumitem}
\usepackage{subcaption}

\usepackage{comment}

\usepackage[inkscapeformat=png]{svg}

\title{GRASP: A Disagreement Analysis Framework to Assess\\ Group Associations in Perspectives}

\author{
   Vinodkumar Prabhakaran* \\
   Google Research\\
   \small{\texttt{vinodkpg@google.com}} \\
   \And
   Christopher M. Homan* \\
   Google Research\\
   \small{\texttt{homanc@google.com}} \\
   \And
  Lora Aroyo* \\
  Google Research\\
  \small{\texttt{l.m.aroyo@gmail.com}} \\
   \AND
   Aida Mostafazadeh Davani \\
   Google Research\\
   \small{\texttt{aidamd@google.com}} \\   
   \And
   Alicia Parrish \\
   Google Research\\
   \small{\texttt{aliciaparrish@google.com}} \\
   \And
   Alex Taylor \\
   Google Research\\
   \small{\texttt{alxtyl@google.com}} \\
   \AND
   Mark Díaz \\
   Google Research\\
   \small{\texttt{markdiaz@google.com}} \\
   \And
   Ding Wang \\
   Google Research\\
   \small{\texttt{drdw@google.com}} \\
   \And
   Gregory Serapio-García \\
   University of Cambridge\\
   \small{\texttt{gs639@cam.ac.uk}} \\
}

\newlist{todolist}{itemize}{2}
\setlist[todolist]{label=$\square$}
\usepackage{pifont}

\usepackage[normalem]{ulem}

\begin{document}

\maketitle

\begin{abstract}

Human annotation plays a core role in machine learning --- annotations for supervised models, safety guardrails for generative models, and human feedback for reinforcement learning, to cite a few avenues. 
However, the fact that many of these human annotations are inherently subjective is often overlooked. 
Recent work has demonstrated that ignoring rater subjectivity (typically resulting in rater disagreement) is problematic within specific tasks and for specific subgroups. Generalizable methods to harness rater disagreement and thus understand the socio-cultural leanings of subjective tasks remain elusive.
In this paper, 
we propose GRASP, a comprehensive disagreement analysis framework to measure \textit{group association} in perspectives among different rater subgroups, and demonstrate its utility in assessing the extent of systematic disagreements in two datasets: (1) safety annotations of human-chatbot conversations, and (2) offensiveness annotations of social media posts, both annotated by diverse rater pools across different socio-demographic axes.
Our framework (based on disagreement metrics) reveals specific rater groups that have significantly different perspectives than others on certain tasks, and helps identify demographic axes that are crucial to consider in specific task contexts.


\end{abstract}

\section{Introduction}

Automatic detection of unsafe, offensive or toxic text has long been an active area of research in Natural Language Processing (NLP). Originally aimed at online content moderation \cite{wulczyn2017ex,founta2018large}, and recently, triggered by academic and governmental calls for action \cite{european2020digital,wh2023},
these efforts are also addressing the urgent need to equip generative technologies with safety guardrails that prevent inadvertent generation of offensive or harmful content \cite{bai2022training,glaese2022improving}.

Much of this work relies on human annotation for evaluating and training offensiveness or safety classifiers, or fine-tuning generative models.
Current approaches largely overlook cultural and individual factors that shape raters' perspectives on what is safe or offensive \cite{aroyo2015truth,talat2016you, salminen2019online,uma2021learning}.
Systematic rater disagreements are instead circumvented by enforcing a single ground truth or using majority vote, which inadvertently marginalizes 
minority perspectives and further amplifies societal biases in data \cite{prabhakaran2021releasing}.

Recent work points to the need for greater diversity in rater pools \cite{thoppilan2022lamda} and proposes ways to incorporate disagreements in the learning pipeline (e.g., \cite{davani2022dealing}). However, incorporating rater diversity at scale is still a challenge, as there are numerous diversity axes to consider, and it is unclear which ones are relevant for particular tasks. 
For instance, 
in sentiment analysis, \citet{prabhakaran2021releasing} found that, while there were systematic disagreements between raters from different racial groups, there were no significant differences across gender groups. In contrast, \citet{Homan2023IntersectionalityIC} found that safety annotations did not differ substantially across race/ethnicity or gender groups individually, but they did differ across intersectional race/ethnicity--gender groups. The lack of effective metrics that can capture such inter-group and intra-group cohesion at scale to determine group-level associations, is a critical issue.

In this paper, we propose GRASP (\underline{Gr}oup \underline{As}sociations in \underline{P}erspctives), a framework to measure the magnitude and strength of systematic diversity of perspectives among rater subgroups. GRASP combines \textit{a suite of metrics} that measure group associations in annotations with \textit{a permutation tests based significance testing approach} that assesses the reliability of these associations without any independence assumptions.
We apply GRASP to two datasets: DICES-350 \cite{Aroyo2023DICESDD}---350 chatbot conversations annotated for safety by 104 raters from a diverse pool across age, gender, and race; and D3 \cite{davani2023disentangling}---social media comments annotated for offensiveness by 4000 raters balanced across cultural regions, gender, and age. GRASP reveals systematic disagreements in annotations along demographic lines,
and shows that it picks up task-dependent group associations in an efficient and effective manner, furthering the objective of identifying meaningful diversity in annotator perspectives in subjective tasks. 

\section{Related work}

Prior work on detecting harmful language, such as toxicity \cite{pavlopoulos2020toxicity,xenos2022toxicity}, offensiveness \cite{davidson2017automated}, and hate speech \cite{warner2012detecting,waseem2016hateful}, has led to curating datasets and developing models for social media content moderation \cite{wulczyn2017ex,founta2018large,vidgen2019challenges}. Recent advancements in conversational AI also increased attention to ensure safety and mitigate potential harms  \cite[e.g.,][]{solaiman2021process,xu2021recipes,shelby2022identifying,si2022toxic,bian2023drop,huang2023chatgpt,santurkar2023whose}.
The latest generation of AI-driven language technologies \cite{openai2023chatgpt,google2022palm,google2023palm2,taori2023alpaca} is based on large language models \cite{openai2023gpt4,touvron2023llama} using reinforcement learning from human feedback (RLHF) \cite{christiano2023deep,ouyang2022training}. Studies show that on human alignment tasks,
rater disagreement can be as high as 40\% \cite{ziegler2020finetuning}. However, not much work has gone into developing scalable methods to meaningfully measure and tackle these high levels of rater disagreement.



Rater disagreement has a long history in NLP research as a challenge for crowd-sourced annotations and as a potential indication of human biases \cite{arhin2021groundtruth,mathew2021hatexplain,sahoo2022detecting,wich2020graph}. Though traditionally viewed as a mark of poor quality data, disagreement is increasingly seen as an important qualitative signal in its own right, one that is present in most tasks that requires human judgement \cite{aroyo2013crowd,hovy2013learning,plank2014linguistically,Klenner2020,basile2021need,weerasooriya-etal-2023-subjective}. 

Empirical analyses of inter-rater disagreements put forth raters’ backgrounds and experiences as crucial to their annotations in such tasks, leading to systematic disagreements \cite[e.g.,][]{prabhakaran2021releasing,kumar2021designing,denton2021whose,sap-etal-2022-annotators,biester-etal-2022-analyzing,deng-etal-2023-annotate,Homan2023IntersectionalityIC,pei-jurgens-2023-annotator}. For instance, raters' demographics, including first language, age, and education, can significantly impact the performance of hate speech and abusive language detectors trained on that rater's behavior \cite{al2020identifying}, and raters' stereotypes about different social groups and attitudes toward racism impact their annotations of hate speech and racist language targeting those groups \cite{sap-etal-2022-annotators,davani2023hate}. Similarly, \citet{davani2023disentangling} show that annotations of offensiveness vary across geo-cultural contexts.  

Consequently, a large body of work has emerged to quantify, model, and measure rater disagreement \cite[e.g.,][]{kairam2016parting,founta2018large,geva2019modeling,chung2019efficient,Obermeyer2019,Liu2019HCOMP,Weerasooriya2020,uma2021learning,weerasooriya-etal-2023-disagreement}. In early work, \citet{hovy2013learning} introduce MACE, an unsupervised item-response model to capture raters' relative trustworthiness to more accurately aggregate annotations into a final label. Recently, \citet{Weerasooriya2020} proposed predictive models for rater disagreement that take into account sampling error, a common problem 
in datasets with very few annotations per item.

Novel modeling efforts have further incorporated raters' demographics and other background attributes to improve the predictions \cite{hovy2015demographic,garten2019incorporating,hovy2021importance}.
Using multi-task modeling frameworks,  \citet{fornaciari2021beyond} add an auxiliary task to predict the soft label distribution over rater labels, \citet{davani2022dealing} model individual raters using a shared network to preserve their systematic disagreements until prediction, and \citet{orlikowski-etal-2023-ecological} incorporates a group-specific layer to assess the benefits of socio-demographic attributes in modeling annotations.
%
\citet{hung2022can} demonstrating the performance improvement when predicting raters' age and gender is coupled with language modeling objectives. Our work provides a framework that anchors on intra-group and inter-group cohesion to qualify the strength of disagreements within and across groups, and provide statistical tests to assess the reliability of these observed group-level patterns.






\section{Group Associations in Annotations}

As outline above, recent studies have established the need to account for systematic rater disagreement in subjective tasks
by demonstrating socio-demographic differences in rater perceptions. However, systematic approaches to reliably assess \textit{whether} and \textit{how much} diversity axes impact disagreement for different tasks are still missing. To address this gap, we introduce a comprehensive analysis framework to measure statistically significant group associations within human annotations.

\subsection{Terminology}

Let us represent a human-annotated dataset as a collection of \emph{items} $\mathbf{X}$ with a corresponding collection of \emph{annotations} $\mathbf{Y}$, obtained from a collection of \emph{raters} $\mathbf{Z}$. Each row $\mathbf{X}_i$ is an item that is annotated, and each corresponding $\mathbf{Y}_i$ captures the annotations for $\mathbf{X}_i$. The columns in $\mathbf{Y}_i$ correspond to individual raters' annotations. In other words, $\mathbf{Y}_{ij}$ represent annotations by rater $\emph{j} \in \mathbf{Z}$ for item \emph{i}.\footnote{Note that $\mathbf{Y}_{ij}$ may be a sparse matrix if each item is labeled by only a handful of raters (which is often the case).}  In its simplest case, $\mathbf{Y}_{ij}$ can be a binary value, but it can be conceived as a vector capturing \emph{j}'s responses to different questions pertaining to \textit{i}, or a one-hot encoding of \emph{j}'s annotation in case of categorical values. Each row $\mathbf{Z}_k$ represents a rater $k$ and the columns of $\mathbf{Z}_k$ contain \emph{group attributes} (e.g., demographic characteristics such as gender, race/ethnicity, and/or age associated with $k$). Let $\Pi$ denote a set of demographic properties, e.g., $\Pi = \{$gender = MALE, age = GenZ$\}$. Then, let  $\mathbf{Z}[\Pi] \subseteq \mathbf{Z}$ denote the subpopulation of raters satisfying that property, and let $\mathbf{Y}_{\mathbf{Z}[\Pi]}$ denote the submatrix of $\mathbf{Y}$ that captures the annotations of that subpopulation of raters according to $\Pi$.

\subsection{Disagreement Analysis Framework}
\label{sec:framework}


We aim to determine whether certain rater groups, defined in terms of their demographic attributes, systematically (and in statistically significant ways) differ from others in terms of their annotations for a given task. For this, we need to measure the (dis)agreement between raters within the group, as well as with those from outside the group. 

\paragraph{In-group Cohesion ($C_I(Y)$)} captures how much cohesion a particular rater group has among themselves. Formally, an \textit{in-group cohesion} metric is a mapping $C_I: 2^{\mathbf{Y}} \rightarrow \mathbb{R}$ where, for any subgroup of annotations $Y \subseteq \mathbf{Y}$, higher values of $C_I(Y)$ indicate higher levels of agreement among $Y$. 
We are interested in $C_I(\mathbf{Y}_{\mathbf{Z}[\Pi]})$, the in-group cohesion among raters who satisfy the set of demographic properties $\Pi$. 

\paragraph{Cross-group Cohesion ($C_X(Y,Y')$)} captures how much one rater group agrees with another rater group. Formally, a \textit{cross-group cohesion} metric is a mapping $C_X: 2^{\mathbf{Y}} \times 2^{\mathbf{Y}} \rightarrow \mathbb{R}$ where, for any pair of subgroups of annotations $Y, Y' \subseteq \mathbf{Y}$, higher values of $C_X(Y, Y')$ indicate higher levels of agreement between the annotations in $Y$ and $Y'$. 
While cross-group cohesion could be calculated for any two given subsets of annotations, we are primarily interested in $C_X(\mathbf{Y}_{\mathbf{Z}[\Pi]},\mathbf{Y}_{\mathbf{Z}[\neg\Pi]})$, the cross-group cohesion between raters satisfying demographic properties $\Pi$ and those who do not.

\paragraph{Group Association Index (GAI):} Both \textit{in-group} and \textit{cross-group cohesion} are useful for assessing the strength of annotation patterns found in a demographic grouping $\Pi$. For instance, high in-group cohesion within $\mathbf{Z}[\Pi]$ and cross-group cohesion between $\mathbf{Z}[\Pi]$ and $\mathbf{Z}[\neg\Pi]$ might just mean that the task has high agreement across the board. On the other hand, $\mathbf{Z}[\Pi]$ having both low in-group and cross-group cohesion might suggest that the raters in general have a hard time agreeing with one another, regardless of the specific grouping $\Pi$. Inspired by graph-theoretic metrics for community detection in networks, such as \emph{modularity} \cite{newman2006modularity}, we introduce a group association index that combines these two aspects into a single score:
\[GAI(\Pi) = \frac{C_I(\mathbf{Y}_{\mathbf{Z}[\Pi]})}{C_X(\mathbf{Y}_{\mathbf{Z}[\Pi]},\mathbf{Y}_{\mathbf{Z}[\neg\Pi]})}\]
The baseline value of $GAI$ is 1; i.e., when $C_I$ and $C_X$ are more or less the same, regardless of their magnitudes, the task annotation patterns have minimal or no group association with $\Pi$. When $C_I$ is larger than $C_X$, the $GAI$ values will be higher than 1, suggesting higher group association with $\Pi$ for the task. On the other hand, for $GAI$ values less than 1, raters agree more with raters outside the group than within the group, suggesting that there are potential patterns of systematic disagreement that are not captured by $\Pi$.


\paragraph{Diversity Sensitivity Index (DSI):} GAI indicates which groups significantly differ from others. There are numerous demographic axes (e.g., gender, age, race/ethnicity, sexual orientation, etc.) along which a rater pool can be diversified. When recruiting raters, which (if any) of these should be prioritized?  It helps to know whether and by how much the subgroups within any axis have a significant $GAI$. This is more insightful than the average $GAI$ value. Hence, we define \textit{diversity sensitivity index} of a task w.r.t. a demographic axis with $K$ groups as the $\max$ of $GAI(\Pi_k$) for $k \in [1,K]$. 
Note that the statistical significance (see below) of the $GAI$ value applies to the $DSI$ value too; i.e., if the $GAI$ value is not significant, the $DSI$ is not either, and vice versa.

\subsection{Significance Testing}
\label{sec:significance}
To ensure our diversity measurements are reliable, it is important to test their significance.
Commonly used tests assume the data items are independently sampled, which doesn't hold in our case, since each annotation depends on all items with the same rater and all raters who annotated that item. 
So we use \emph{permutation tests} to control for these dependencies.

\paragraph{Null hypothesis:}
For any in-group cohesion (or cross-group divergence) metric $C_I$ (or $C_X$), our null hypothesis $\mathbf{H_0}$ is
\begin{quote}
$\mathbf{H_0}$: Value of $C_I$ (or $C_X$) for any (pair of) subpopulation(s) $\mathbf{Y }_{ \mathbf{Z}[\Pi_1]}$ (, $\mathbf{Y }_{\mathbf{Z}[\Pi_2]}$) is independent of demographic profile(s) of member(s) of $\Pi_1$ (and $\Pi_2$).
\end{quote}
To test $H_0$, we randomly shuffle the raters demographic profiles, measure the test statistic after each shuffle, and then count how many times the shuffled statistic exceeds the observed value. If the observed value is significant, then \emph{\textbf{only a small percentage 
of the measurements from random groups should exceed the observed value}}.
%
Formally, p-value 
of $C_I$ is defined as:

\begin{multline*} \footnotesize
p_{C_I}(\mathbf{Y}_{\mathbf{Z}[\Pi_1]}) =_{\rm def}\\
\left\{
    \begin{array}{l}
    \|\{s_i^* : s_i^* < C(\mathbf{Y}_{\mathbf{Z}[\Pi_1]})\}\|/N \\
   \hspace{1.5cm} \mbox{if } C(\mathbf{Y}_{\mathbf{Z}[\Pi_1]}) < s_{\lfloor N/2 \rfloor}^*,  \vspace{.2cm}\\
    \|\{s_i^* : s_i^* > C(\mathbf{Y}_{\mathbf{Z}[\Pi_1]})\}\|/N \\
    \hspace{1.5cm} \mbox{otherwise.}
    \end{array}\right.
\end{multline*}

\noindent where $N$ is a large number and $s_1^*, \ldots, s_N^*$ are computed by the following pseudocode:

\vspace{4pt}
\begin{algorithmic}
\STATE $i \gets 0$
\WHILE{$i < N$}
    \STATE $\mathbf{Z}^* \gets$ randomly permute the rows of $\mathbf{Z}$ (but fix the indices, so that the rows map to the same annotations even though their demographics have changed)
    \STATE $i \gets i + 1$
    \STATE $s_i^* \gets C(\mathbf{Y}_{\mathbf{Z}^*[\Pi_1]})$
\ENDWHILE
\STATE reorder $s_1^*, \ldots, s_N^*$ in ascending order.
\end{algorithmic}

\vspace{4pt}

\noindent The p-value $p_{C_X}(\mathbf{Y}_{\mathbf{Z}[\Pi_1]}, \mathbf{Y}_{\mathbf{Z}[\Pi_2]})$ of $C_X$ is defined as above, except that we replace $C_I(\mathbf{Y}_{\mathbf{Z}[\Pi_1]})$ with
$C_X(\mathbf{Y}_{\mathbf{Z}[\Pi_1]}, \mathbf{Y}_{\mathbf{Z}[\Pi_2]})$ (and $C_I(\mathbf{Y}_{\mathbf{Z}^*[\Pi_1]})$ with
$C_X(\mathbf{Y}_{\mathbf{Z}^*[\Pi_1]}, \mathbf{Y}_{\mathbf{Z}^*[\Pi_2]})$).

\paragraph{Multiple test correction:}
If numerous tests are conducted and the null hypothesis is true, then by the Law of Large Numbers 
some of them are likely to have small p-values, making them falsely appear to be significant (type I error). 
There is no widely accepted best practice for dealing with this problem. Some researchers advocate never using p-values for exploratory research \cite{hak2014after,traf:ed:2015} or to apply corrections such as Bonferonni \cite{bonferroni1936teoria,holm1979simple} against the family-wise error rate. Other researchers see those approaches as too restrictive, which can lead to important discoveries being missed \cite{gaus2015interpretation,goeman2011mult,rubin2017p}. 
%
We adopt a mixed approach and report two levels of significance: significance with no correction whatsoever and with Benjamini-Hochberg \emph{false discovery rate (FDR)} correction \cite{benjamini1995controlling}. 
%

\subsection{Metrics}
\label{sec:metrics}

The concepts introduced in \S\ref{sec:framework} are \textit{metric-agnostic}, and the choice of metric must be justified for each experiment.
Here, we describe the three kinds of metrics we use in this paper for both $C_I$ and $C_X$; we compare and contrast what these metrics are sensitive to and what they reveal. 




\subsubsection{In-group Cohesion Metrics}
\label{sec:metrics_ic}


\paragraph{IRR:}
We use {IRR} (Inter-rater reliability, particularly, Krippendorff's alpha \citealt{krippendorff2004reliability}) to measure within-group agreement while controlling for class imbalance. Krippendorff's alpha has an advantage over other IRR metrics: it can handle an arbitrary number of raters, answer options and items at one time, and it unifies and generalizes a number of other IRR metrics, including Scott's pi and Fleiss' kappa \cite{krippendorff2004reliability}. 
It is formulated as $1 - \frac{o_d}{e_d}$, where $o_d$ is the mean observed disagreement between pairs of distinct raters, and $e_d$ is the class-imbalance-controlling term. The $o_d$ term is, effectively, hamming distance and $e_d$ is the expected amount of disagreement, under the assumption that each rater's responses are randomly distributed among the conversations they label (but each rater's marginal distribution of annotations is fixed), independent of the other raters' responses. 
    

    
\paragraph{Plurality size:} IRR and our many other metrics are based on counting the (dis)agreements between pairs of raters. But in practice, raters are often seen as populations whose annotations are taken as \emph{votes}, where the most popular annotation (i.e., majority vote) becomes the gold standard response.  Thus, a very natural measurement of agreement is the fraction of raters who belong to the most popular choice (similar to \citealt{prabhakaran2021releasing}'s approach). This metric is less sensitive to class imbalance than metrics that count pairwise disagreements. It is computed by iterating over each item, taking the argmax over the distribution of responses, and then taking its mean over all pairs.

    
\paragraph{Negentropy:} IRR measures pairwise agreement between raters and plurality size captures the impact of disagreement in the rating aggregation process. Another common way to measure disagreement in groups, used in polls and surveys, is to estimate the distribution of annotations associated with each item. Entropy is a common metric for measuring the randomness of a probability distribution, such as the annotations from multiple raters to a safety question about a conversation. It captures how evenly distributed the ranges of responses are. 
To orient all our metrics so that larger numbers mean more agreement, we report \emph{negentropy} \cite{brillouin1953negentropy}: for each conversation, we compute the entropy over the distribution of responses. Then we subtract this from the maximum value entropy can take over the response domain. For a domain with $n$ possible responses, this number is $\ln n$. Finally, we take the mean over all conversations.


\subsubsection{Cross-group Divergence Metrics}
\label{sec:metrics_cd}

Analogous to our in-group cohesion metrics, we focus on three cross-group cohesion metrics. 

\paragraph{XRR:} \emph{Cross-replication reliability} \cite{wong2021cross} is similar to Krippendorff's alpha, except that the pairs of raters being compared come from separate groups. Like alpha, XRR can handle arbitrary numbers of raters, answer options and items. And it also controls for class imbalance. 

\paragraph{Voting agreement:} For across-group agreement, it is equally natural, by analogy to plurality size, to compare two groups as if they were voting blocks. For each item, we compute the plurality choice for each group. To account for class imbalance, we compute Krippendorff's alpha over all conversations between the two groups, based on each group's plurality choices. 
Although straightforward, we are not aware of this method proposed as a group-level divergence metric.


\paragraph{Cross-negentropy:} Cross-entropy is algorithmically similar to entropy but is computed over two distributions, not one. We define cross-negentropy in an analogous manner to negentropy. 


\section{Experiments}
\label{sec:experiments}

\subsection{Data}
\label{sec:data}

We apply our metrics to the two datasets: \textbf{DICES-350} \cite{Aroyo2023DICESDD},\footnote{https://github.com/google-research-datasets/dices-dataset/tree/main/350} and \textbf{D3} \cite{davani2023disentangling}. The DICES-350 dataset is a curated sample of 8k multi-turn conversation corpus generated by human agents interacting with a generative AI-chatbot \cite{thoppilan2022lamda} in an adversarial setting. These conversations were then annotated for safety by a diverse rater pool.
The D3 dataset contains a curated sample of social media posts from Jigsaw datasets \cite{Jigsaw-bias, Jigsaw-toxic}, annotated for offensiveness in text. 
We choose the DICES-350 and D3 datasets as they both contain fully replicated annotations from a diverse rater pool along with their demographic details, enabling our in-depth and fine-grained group-level analyses.

\begin{table}
\small
\centering
\scalebox{.9}{
\begin{tabular}{llp{7ex}p{8ex}p{10ex}}
\toprule
\textbf{Dataset} &
 \textbf{Items} &
 \textbf{Rater pool} &
 \textbf{Raters per item} &
 \textbf{Total \mbox{annotations}} \\
\midrule

DICES-350 & 350 & 104 & 104 
&582,400\\
D3 & 4554 & 4309 & 24 & 150,702 \\
\bottomrule
\end{tabular}}
\vspace{1ex}
\caption{\label{tab:datasets-overview1} DICES-350 and D3 dataset annotation stats.}
\end{table}

\paragraph{DICES-350} 
contains annotations for safety along 16 dimensions for all 350 conversation by 123 unique raters based in the US.
The authors of DICES-350 aimed for
an approximately equal numbers of raters in each of the 12 demographic groups (3 x 4 design) created by fully crossing age groups (GenZ, Millennial, GenX+) with race/ethnicity (Asian; Black; Latine/x; White).  
All raters annotated all 350 conversations. 
We limit our study to 104 raters after removing 19 raters who were deemed unreliable by the authors of DICES-350.
See Table~\ref{tab:datasets-dices-overview} for breakdowns of the demographic groupings along race, gender, and age.

\begin{table}[t!]
\small
\centering
\begin{tabular}{@{}ccccccc@{}}\toprule
\multicolumn{5}{c}{DICES-350}\\\midrule
    \multirow{2}{*}{Race} & \multicolumn{2}{c}{Gender} & \multicolumn{3}{c}{Age} \\\cmidrule(r){2-3}\cmidrule(r){4-6}
    & F & M & GenZ & Mill. & GenX+\\\midrule
    As. & 9 & 12 & 4 & 12 & 5 \\
    Bl. & 16 & 7 & 13 & 5 & 5 \\
    Lat. & 12 & 10 & 6 & 7 & 9 \\
    Multi. & 4 & 9 & 6 & 2 & 5 \\
    Wh. & 16 & 9 & 5 & 2 & 18 \\
    \bottomrule
\end{tabular}\\
\caption{\label{tab:datasets-dices-overview} DICES-350 raters in various demographic intersectional groups. Race/ethnicity information is abbreviated for space: Bl: Black; Wh: White; As: Asian; Lat: Latine; Multi: Multi-racial.}
\end{table}

\begin{table}[t!]
\small
\centering
\begin{tabular}{@{}cccccccc@{}}\toprule
\multicolumn{6}{c}{D3}\\\midrule
    \multirow{2}{*}{Region} & \multicolumn{3}{c}{Gender} & \multicolumn{3}{c}{Age} \\\cmidrule(r){2-4}\cmidrule(r){5-7}
    & F & M & O & 18-30 & 30-50 & 50+\\\midrule
    AC. & 205 & 306 & 5 & 269 & 168 & 79\\
    ICS. & 245 & 308 & 1 & 237 & 198 & 119\\
    LA. & 275 & 271 & 3 & 302 & 176 & 71\\
    NA. & 325 & 220 & 6 & 263 & 175 & 113\\
    Oc. & 307 & 203 & 7 & 161 & 221 & 135\\
    Si. & 249 & 280 & 11 & 208 & 228 & 104\\
    SSA. & 219 & 309 & 2 & 320 & 157 & 53\\
    WE. & 294 & 252 & 6 & 259 & 172 & 121\\\bottomrule
\end{tabular}\\
\caption{\label{tab:datasets-d3-overview} D3 dataset raters in various intersectional groups. Region names abbreviated for space: AC: Arab Culture; ICS: Indian Cultural Sphere; LA: Latin America; NA: North America, Oc: Oceania, Si: Sinosphere; SSA: Sub-Saharan Africa, WE: Western Europe.}
\end{table}

The safety annotation dimensions 
cover a variety of safety violations, including harmful content, unfair bias, misinformation, and political endorsements,
and raters may respond \emph{Safe}, \emph{Unsafe}, or \emph{Unsure}. 
%
We compute a single safety response for each rater-conversation pair by aggregating the responses into a single, overall safety response. 
For any conversation, if \textit{any} of the safety annotations is \emph{Unsafe}, then we label the entire conversation as unsafe. Otherwise, if any of the safety annotations is \emph{Unsure}, then so is the aggregated response. Otherwise, the aggregated response is \emph{Safe}. 
In other words, it only takes one reason for a conversation to be unsafe and, conversely, if a conversation is unsafe, it need only be unsafe for one reason.

\paragraph{D3} is similarly annotated by a diverse pool of 4k raters across 8 geo-cultural regions and 21 countries. Each item in the dataset was annotated by at least three raters in each region ($\sim$24 annotations per item). The annotation effort aimed for capturing an approximately equal number of raters ($\sim$450) from each region and equal ratio of representation for various demographic group across age (18 to 30, 30 to 50, and more than 50 years old) and genders (Man, Woman, and Other). 
See Table~\ref{tab:datasets-d3-overview} for the breakdown of the demographic groups across different regions, gender, and age groups. 

Raters were asked to label the textual items' level of offensiveness on a 5-point Likert scale, 1 being \textit{not offensive at all} and 5 being
\textit{extremely offensive}, with the option of choosing \textit{Unsure}. We treated a score of 3 or higher as being \textit{Offensive}, in line with the dataset creators \cite{davani2023disentangling}.


\begin{table}[h!]

\centering
\scalebox{.75}{
\begin{tabular}{lllll}
\toprule
\multicolumn{5}{c}{\textbf{DICES-350}}\\\midrule
Dimension &               Group &                  IRR &                  XRR &            GAI \\
\midrule
age &              gen x+ &               $\downarrow$0.166 &               $\downarrow$0.171 &        $\downarrow$0.975 \\
age &               gen z &               $\downarrow$0.166 &               $\downarrow$0.172 &         $\downarrow$0.966 \\
age &           millenial &                 $\uparrow$0.189 &                 $\uparrow$0.179 &                                 $\uparrow$1.052 \\
\midrule
gender &             Man &                 $\uparrow$0.187 &                 $\uparrow$0.175 &                                $\uparrow$1.071 \\
gender &            Woman &               $\downarrow$0.160 &                 $\uparrow$0.175 &                           $\downarrow$0.916 \\
\midrule
race &               As. &               $\downarrow$0.145 &               $\downarrow$0.166 &                            $\downarrow$0.872 \\
race &               Bl. &                 $\uparrow$0.193 &                 $\uparrow$0.181 &                              $\uparrow$1.063 \\
race &              Lat. &    \textbf{$\uparrow$0.215*} &    \textbf{$\uparrow$0.189*} &          \textbf{$\uparrow$1.139*} \\
race &         Multi. &               $\downarrow$0.153 &               $\downarrow$0.168 &                $\downarrow$0.916 \\
race &               Wh. &               $\downarrow$0.145 &  \textbf{$\downarrow$0.159*} &                  $\downarrow$0.908 \\
\midrule
\multicolumn{5}{c}{Statistically Significant Intersections}\\\midrule
race, gender &        As., Woman &  \textbf{$\downarrow$0.073*} &  \textbf{$\downarrow$0.134*} &   \textbf{$\downarrow$0.540*} \\
race, gender &        Bl., Woman &    \textbf{$\uparrow$0.213*} &                 $\uparrow$0.188 &          \textbf{$\uparrow$1.130*} \\
race, gender &       Lat., Woman &    \textbf{$\uparrow$0.238*} &   \textbf{$\uparrow$0.199**} &        \textbf{$\uparrow$1.196*} \\
race, gender &          Wh., Man &    \textbf{$\uparrow$0.218*} &               $\downarrow$0.173 &    \textbf{$\uparrow$1.262**} \\
race, gender &        Wh., Woman &  \textbf{$\downarrow$0.114*} &  \textbf{$\downarrow$0.152*} &       \textbf{$\downarrow$0.752*} \\
\end{tabular}}

\scalebox{.7}{
\begin{tabular}{lllll}
\toprule
\multicolumn{5}{c}{\textbf{D3}}\\\midrule
Dimension & Group &  IRR &  XRR & GAI \\
  \midrule
age&  (18,30)
&\textbf{$\uparrow$0.115**}
&$\uparrow$0.107
&\textbf{$\uparrow$1.068**}
\\
age&  (30,50)
&\textbf{$\downarrow$0.089**}
&$\downarrow$0.104
&\textbf{$\downarrow$0.850**}
\\
age&  50+
&$\uparrow$0.110
&$\uparrow$0.111
&$\uparrow$0.999
\\\midrule
gender&  Woman
&$\uparrow$0.110
&$\uparrow$0.108
&$\uparrow$1.024\\
gender&  Man
&$\downarrow$0.105
&$\uparrow$0.107
&$\downarrow$0.976
\\
gender&  Other
&$\uparrow$0.209
&$\downarrow$0.096
&\textbf{$\uparrow$2.172*}
\\\midrule
region&  AC.
&\textbf{$\uparrow$0.133**}
&$\uparrow$0.113
&\textbf{$\uparrow$1.174*}
\\
region&  ICS.
&$\downarrow$0.103
&\textbf{$\downarrow$0.099*}
&$\uparrow$1.043
\\
region&  LA.
&\textbf{$\uparrow$0.129**}
&$\uparrow$0.112
&\textbf{$\uparrow$1.152*}
\\
region&  NA.
&\textbf{$\uparrow$0.143**}
&$\uparrow$0.110
&\textbf{$\uparrow$1.307**}
\\
region&  Oc.
&$\uparrow$0.118
&$\downarrow$0.103
&\textbf{$\uparrow$1.145*}
\\
region&  Si.
&\textbf{$\downarrow$0.087*}
&\textbf{$\downarrow$0.087**}
&$\downarrow$1.002
\\
region&  SSA.
&\textbf{$\uparrow$0.142**}
&$\downarrow$0.104
&\textbf{$\uparrow$1.361**}
\\
region&  WE.
&\textbf{$\uparrow$0.135**}
&$\uparrow$0.111
&\textbf{$\uparrow$1.222**}
\\
\midrule
\multicolumn{5}{c}{Statistically Significant Intersections}\\\midrule
region, age& ICS., (18,30)
&\textbf{$\downarrow$0.063**}
&$\downarrow$0.100
&\textbf{$\downarrow$0.634*}
\\
region, age& ICS., (30,50)
&\textbf{$\downarrow$0.060*}
&$\downarrow$0.100
&\textbf{$\downarrow$0.601*}
\\
region, gender& ICS., Woman
&\textbf{$\downarrow$0.070*}
&$\downarrow$0.106
&\textbf{$\downarrow$0.655*}
\\
region, age& LA., (18,30)
&\textbf{$\uparrow$0.143**}
&$\uparrow$0.118
&\textbf{$\uparrow$1.216*}
\\
region, gender& LA., Woman
&\textbf{$\uparrow$0.143**}
&$\uparrow$0.111
&\textbf{$\uparrow$1.290*}
\\
region, gender& NA., Woman
&\textbf{$\uparrow$0.153**}
&$\uparrow$0.116
&\textbf{$\uparrow$1.314**}
\\
region, age& Oc., (30,50)
&$\uparrow$0.112
&\textbf{$\downarrow$0.089**}
&\textbf{$\uparrow$1.255*}
\\
region, gender& Oc., Woman
&\textbf{$\uparrow$0.133*}
&$\uparrow$0.110
&\textbf{$\uparrow$1.208*}
\\
region, age& Si., (30,50)
&\textbf{$\downarrow$0.033**}
&\textbf{$\downarrow$0.082**}
&\textbf{$\downarrow$0.405**}
\\
region, age& Si., 50+
&$\uparrow$0.137
&\textbf{$\downarrow$0.061**}
&\textbf{$\uparrow$2.225**}
\\
region, gender& Si., Woman
&$\downarrow$0.100
&\textbf{$\downarrow$0.081**}
&\textbf{$\uparrow$1.237*}
\\
region, age& SSA., (18,30)
&\textbf{$\uparrow$0.146**}
&$\downarrow$0.107
&\textbf{$\uparrow$1.365**}
\\
region, age& WE., (18,30)
&\textbf{$\uparrow$0.177**}
&\textbf{$\uparrow$0.126**}
&\textbf{$\uparrow$1.402**}
\\
region, gender& WE., Woman
&\textbf{$\uparrow$0.151**}
&$\uparrow$0.118
&\textbf{$\uparrow$1.284*}
\\
\bottomrule
 \end{tabular}}

\caption{Results for in-group and cross-group cohesion, and GAI. Significant results are in \textbf{bold}: * for significance at $p < 0.05$, ** for significance after Benjamini-Hochberg correction.
A $\downarrow$ (or $\uparrow$) means that the result is less (or greater) than expected under the null hypothesis. 
GAI results based on $C_X=$ XRR and $C_I=$ IRR. 
}
\label{tab:results-summary}
\end{table}

\subsection{Results}
\label{sec:results}

We report results of our analysis using IRR and XRR as the in-group and cross-group cohesion metrics in for both DICES-350 and D3 datasets in Table \ref{tab:results-summary}. We focus on IRR and XRR based analysis in this section, but the full results using all other metrics are presented in Tables~\ref{tab:dices-results-all}, ~\ref{tab:d3-results-all}, and \ref{tab:d3-interactions-results}.

We investigate groupings along age, gender, and either race/ethnicity (DICES-350) or region (D3).
For DICES-350, we also explore intersectional groups along race/ethnicity and gender (some of the intersections of age and race/ethnicity are too small to reasonably assess significance), 
while
we explored the intersection of region with both age and gender groups in the D3 dataset. 
Results for all intersections and statistically significant intersections are reported in Tables \ref{tab:dices-results-all}--\ref{tab:d3-interactions-results} and  \ref{tab:results-summary}, respectively.


\paragraph{DICES-350 results:} Only race/ethnicity groupings show significant results on their own, suggesting age and gender doesn't matter. However, looking at intersectional groups, Latine women have the highest in-group cohesion (0.238), followed by White men (0.218), Latine raters (0.215), and Black women (0.213). Asian women have the lowest score (0.073), followed by White women (0.114). Latine women also have the highest cross-group cohesion (0.199),
followed by Latine raters (0.189). Asian women have the lowest score (0.134), followed by White women (0.152) and White raters (0.159).
White men have the highest GAI score (1.262) 
followed by Latine women (1.196), Latine raters (1.139), and Black women (1.130). 
Some groups have GAIs significantly lower than baseline; Asian women have the lowest GAI (0.540), followed by White women (0.752), suggesting that these groups have constituent subgroups that have more agreement with raters outside this group.

The DSI metric looks at what is the highest GAI for each diversity axis (including intersectional axes) we consider. In the DICES-350, we observe the higher DSI for the intersectional axis of gender and race (1.262 for White men), followed by race considered alone (1.139 for Latine raters).
These numbers suggest that it is crucial to prioritize recruiting raters with a diverse representation along race and gender, while diversifying along age may be less crucial based on our results for this task. Note that, although unlikely, applying our framework along other intersectional axes including age may reveal other group associations.

\paragraph{D3 results:} Here, 18-to-30-year-old Western Europeans have the highest IRR (0.177), followed by North American women (0.153) and Western European women (0.151). 
Lowest scores are reported for 30-to-50-year-old raters from Sinosphere (0.033) and Indian Cultural Sphere (0.060),
followed by 18-to-30-year-old (0.063), and women (0.070) groups from Indian Cultural Sphere.  
18-to-30-year-old Western Europeans also have the highest XRR (0.126) 
followed by non-significant scores for Western European women (0.118) and North American women (0.116). Lowest XRR is reported for 50+-year-old raters of Sinosphere (0.061), followed by Sinosphere women (0.081) and 30-to-50-year-old Sinosphere raters (0.082), all significant after BH corrections. 
In terms of GAI scores, 50+-year-old raters of Sinosphere (2.225), and raters identifying with non-binary genders (2.172) report the highest GAI, followed by 18-to-30-year-old groups in Western European (1.402) and Sub-Saharan Africa (1.365); all significant after BH correction. Notably, unlike the DICES-350, different age and region groups have significantly high GAI scores; 18-to-30-year-old (1.068), North America (1.307), Sub Saharan Africa (1.361), and Western Europe (1.222). Interestingly, intersectional results demonstrate that while women in general did not report high GAI, subgroups of women in different regions show more in-group agreement. 

We observe the highest DSI for the intersectional axis of region and age (Sinosphere, 50+) at 2.225, followed by a high DSI for gender (Other) at 2.172. This shows the importance of prioritizing raters from non-binary gender groups and specific subgroups along region and age to capture important diverse perspectives in assessing offense.

\section{Discussion}

We propose the GRASP framework that provides a means to assess the cohesion and strength of group associations along different axes of diversity that matter for a given task, identifying different groups, including intersectional groups, that are relevant for specific tasks. 
GRASP is generic and versatile, however different task and data settings can lead to different results in group associations, depending on the metrics we use. Our intent in presenting this breadth of results using different metrics is both to show the versatility of the framework w.r.t. underlying metrics, and also to demonstrate the differences that different underlying metrics yield. In practice, one should choose the metrics suitable to the specific task and data characteristics (e.g., number of raters, replication factor, and data skew).

\textbf{Task specific insights:} Our analysis provides insights about specific rater groups for each task. For instance, in the conversational safety task (DICES-350), White men having the highest and Asian women the lowest in-group cohesion. Interestingly, White women and Asian men had opposite cohesion trends from their alter-genders. This suggests that men are driving the high cohesion observed in White raters, and that women and men counteract each other in the weak effects observed in Asian raters overall. High coherence among White men is due to their strong tendency to prefer \emph{Safe} to \emph{Unsafe} annotations by a nearly $3:1$ ratio.
On the other hand, for the offense annotation task (D3), most regional groups show significant group associations. Notably, Indian cultural sphere and Sinosphere shows no significant in-group cohesion (or GAI), although 50+ groups within Sinosphere show high in-group cohesion. Age is a notable factor across board, both individually and within intersectional groups, suggesting the need for diversification of rater pools around age groups.

\textbf{Flexibility of group granularity:} Our analysis is generic enough that it can be applied groups defined by any subset of demographic characteristics, enabling it to easily reveal intersectional group associations. 
For instance, although age and gender groups revealed no association for safety, intersectional analysis revealed that gender plays a substantial role in driving race-level group tendencies.

\textbf{Flexibility of metrics}: Our framework is extensible to any (comparable) underlying in-group cohesion and cross-group divergence metrics. We observe that the values across our metrics vary (see Table~\ref{tab:dices-results-all} \& \ref{tab:d3-results-all}); IRR numbers are relatively low (around $0.2$) while other metrics report much higher agreements. These disparities may point to potential overcompensation for class imbalance ($2:1$ for \emph{safe} to \emph{unsafe}) in the IRR metric. IRR is typically used to compare small groups of raters. With larger groups of raters there are quadratically more pairs of raters, and the high dimensionality of the response vectors (350 responses per rater) means that all pairs can potentially be very different from each other: there is both more space to disagree and more disagreements to count. Negentropy and plurality size are less sensitive to these effects, since they are both based on the distributions of all raters, not on the pairwise relationships between all raters. Future work should look into which metrics may be more suitable in specific task and data settings (e.g., number of raters, replication factor, etc.).
 
\textbf{Versatility across dataset characteristics}: The two datasets we applied our framework to differ not only on the underlying tasks, but also on other dataset characteristics/structure. DICES-350 contains fully parallel annotations (i.e., all 104 annotators annotated all 350 items), whereas D3 contains batches of annotations where sets of 35 items contain fully parallel annotations from 24+ raters. These differences did not hinder the applicability of the analysis framework. In fact, the D3 analysis provides a potential pathway where such highly parallel annotations by broadly diverse rater pools could be performed in early phases, that can then inform more streamlined data collection through curated rater pools representing selected diversity axes based on this analysis, essentially saving cost while ensuring diversity in data.

\textbf{Exploratory Analysis:} Our approach also illustrate the usefulness of significance testing to exploratory analysis. 
{We see the role of significance testing in exploratory research as a compass} that provides perspective in light of conflicting results that lack inherent scales for interpretation.
While they impose a hefty computational burden, 
the permutation tests control for joint dependencies in the data between raters and conversations that simpler tests do not. However, we believe the extra computational effort is well worth the trouble, especially in informing rater recruitment decisions. 

\section{Conclusion}
We introduced GRASP, an analytical 
framework to measure systematic 
diversity in annotations among rater subgroups, to better understand the socio-cultural leanings of subjective tasks. We proposed a group association index that combines in-group and cross-group cohesion, along with statistical significance using permutation tests.
Applying this framework to two datasets of subjective annotations,
we demonstrated how it reveals systematic disagreements across various intersectional subgroups. 
Our work contributes to the efforts on bringing in diverse perspectives in data in an efficient and effective manner, furthering the goal of robust socio-technical evaluations of AI models.

Future work could explore adopting our framework to measure systematic disagreements between rater groups in other subjective tasks (e.g., \citealt{kumar-2022-answer}) to further validate its utility. Furthermore, our analysis framework provides actionable insights for practitioners to help prioritize demographic axes when diversifying rater pools. Future work could investigate how the framework may enable dynamic data collection that can adapt to emergent group associations among raters across different types of content and tasks.
While our framework identifies systematic disagreements between groups, further investigation is needed to understand the underlying reasons that cause these disagreements, for instance, individual moral values \cite{davani2023disentangling}.


\section{Limitations}

We acknowledge that the demographic breakdown in both datasets is a simplified representation of the population at large. We assume this was done to facilitate recruitment of raters in each group and to allow for less complexity in analysing intersecting groups. However, our analysis framework was applied on two independent datasets with different rater pools, demographic breakdowns and data collection designs, which points to its generalizability. Provided more granular demographic data, we are confident the frameworks can be readily applied.

We recognize that further research is needed to extend such analysis to other intersectional groups that we have not been investigated in this paper. For example, we believe that further slicing the ethnicity, native languages and age groups is likely to reveal further insights and provide additional evidence of systematic differences between different groupings of raters. Due to page limit this paper focuses on introducing the disagreement analysis framework, and provide initial analysis to demonstrate its utility in revealing significant group associations along socio-demographic lines.

Finally, we recognize more work is needed to distinguish \emph{good} from \emph{bad} disagreement. We focused on revealing statistically significant cohesion within groups (and lack of it across groups), which may weed out noisy disagreements. However, more work is needed to disentangle disagreements that are important to retain in the interest of retaining diverse perspectives, vs. those that are undesirable from a practitioners' perspective (e.g., lack of training in a particular rater platform/pool).

While the use of significance tests in exploratory analysis is controversial \cite{balluerka2005controversy}, there is usually a degree of arbitrariness in their use, for instance, in the choice of level (e.g., $p = 0.05$, in our case), if nothing else. In the case of exploratory research such as ours, one must be careful not to abuse significance testing. For instance, we deliberately held back on a deeper exploration of intersectionality to reduce the risk of p-hacking (see discussion in \S~\ref{sec:significance}, \ref{sec:results}). We also note that we have many more significant results at the $p=0.05$ level than chance would predict. 
There is also arbitrariness in the metrics used. For instance, there isn't uniform agreement on how to interpret well-established metrics such as Krippendorff's alpha. 

\section{Statement of Ethics}
\label{sec:ethics}

Collecting and analyzing socio-demographic information of annotators raise significant ethical considerations.
Hence, all the demographics data present in DICES-350 \cite{Aroyo2023DICESDD} and D3 \cite{davani2023disentangling} datasets are self-declared. Raters were presented a consent form before signing up for both studies to inform them about the gathering of personal demographics and that the content to be rated is adversarial (i.e., would possibly contain offensive content). All demographics questions had the option "Prefer not to answer". All data was collected in an anonymized fashion.
Raters were allowed to quit the study at any time. 
Similar precautions should be taken while building new datasets with socio-demographic information.

\section*{Acknowledgements}
\label{sec:ack}

We thank Chris Welty and Katherine Heller for valuable feedback on early drafts of the manuscript. We also thank anonymous reviewers for their constructive feedback during the peer review process.

\bibliography{anthology,emnlp2023_disagreement}


\clearpage
\appendix

\section{Appendix}
\label{sec:appendix}

\begin{figure*}
    \centering
    \includegraphics[width=\textwidth]{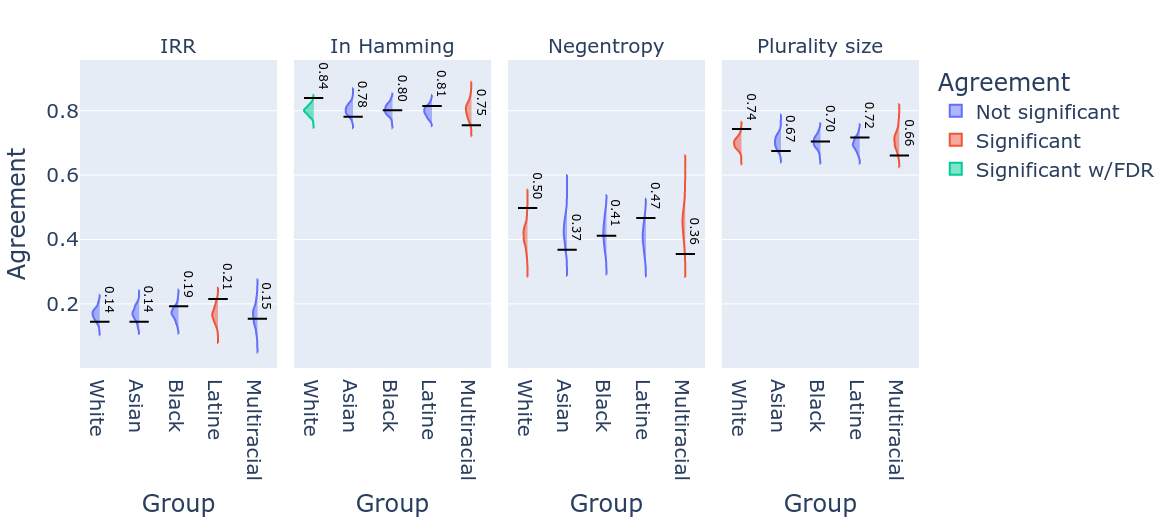}
    \caption{Within-group agreement metrics, by race/ethnicity. Negentropy and plurality size indicate that White raters have significantly more, and Multiracial significantly less, agreement than other races/ethnicities. IRR indicates that Latine raters have significantly more agreement than other races/ethnicities}
    \label{fig:within-group-race}
\end{figure*}

\begin{figure*}
    \centering
    \includegraphics[width=\textwidth]{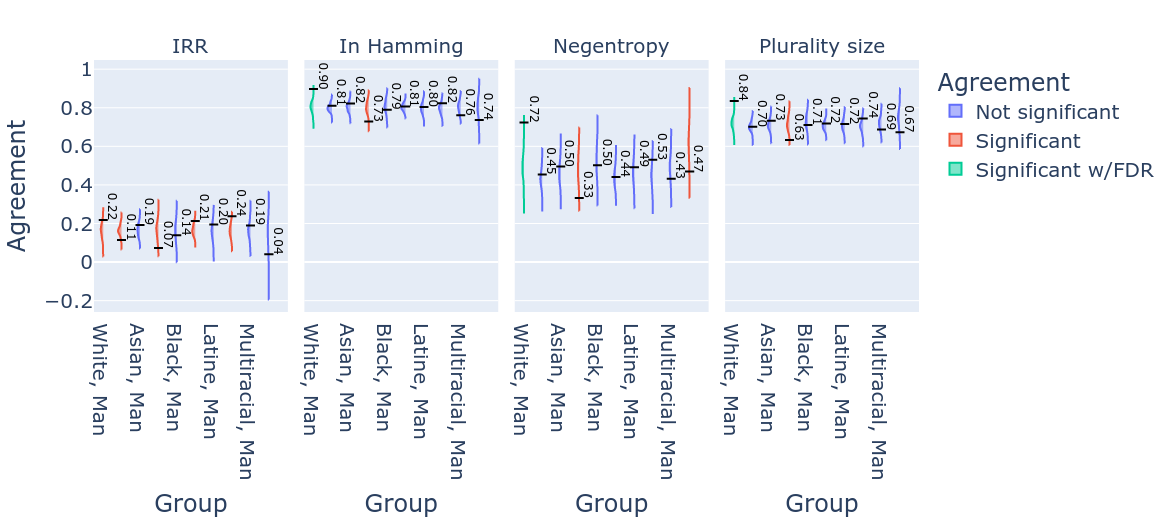}
    \caption{Within-group agreement metrics, by race/ethnicity and gender. Histograms represent the distribution of agreement values under the null hypothesis. Black horizontal bars represent the observed values. These results show that white men have significantly less agreement than other groups, according to negentropy and plurality size, neither of which control for class imbalance. IRR shows that with controlling for class imbalance between \emph{safe} and \emph{unsafe} annotations, the amount of agreement is more moderate. Asian women show nearly the opposite results, with less agreement than other groups unless class imbalance is controlled.}
    \label{fig:within-group-race-gender}
\end{figure*}

\begin{figure*}
    \centering
    \includegraphics[width=\textwidth]{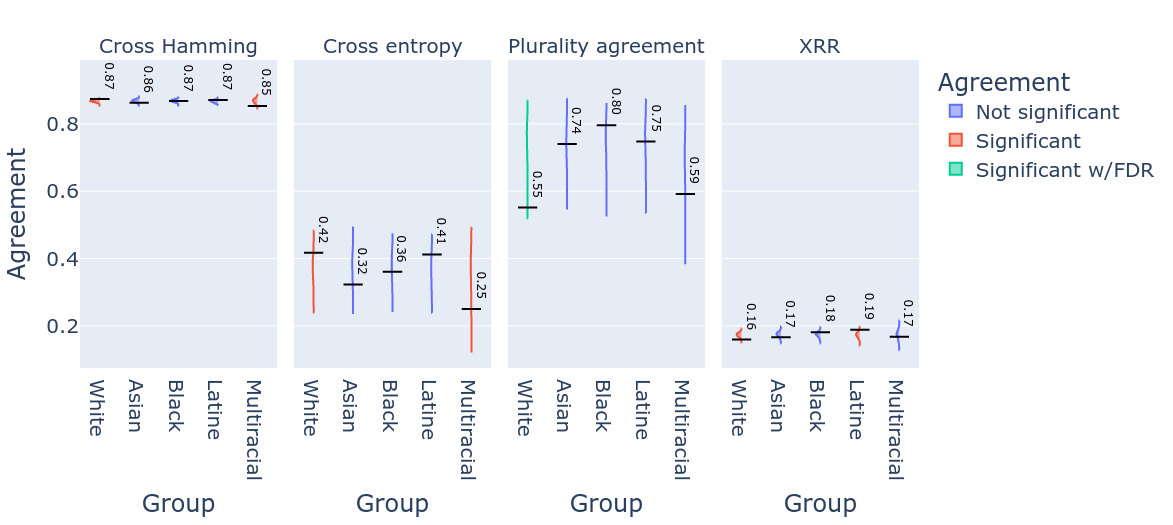}
    \caption{Across-group agreement metrics, by race/ethnicity. Histograms represent the distribution of agreement values under the null hypothesis. Black horizontal bars represent the observed values. White and multiracial voters show less overall agreement with others. Latine voters show more agreement with others.}
    \label{fig:Across-group-race}
\end{figure*}

\begin{figure*}
    \centering
    \includegraphics[width=\textwidth]{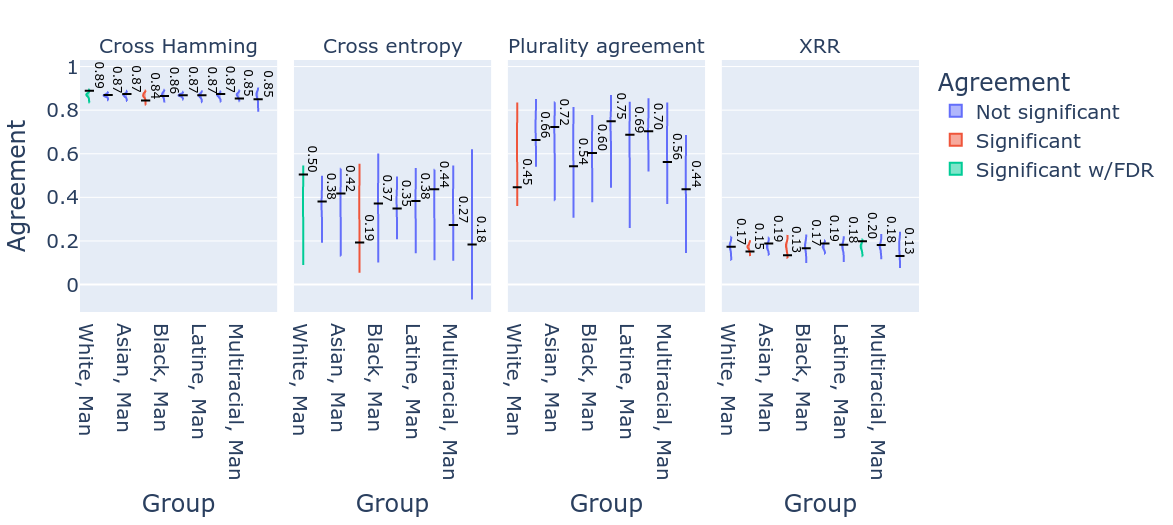}
    \caption{Across-group agreement metrics, by race/ethnicity and gender. Histograms represent the distribution of agreement values under the null hypothesis. Black horizontal bars represent the observed values. Here, white men show signs of significantly low plurality agreement. With other groups. Yet safety agreement is significantly high (though will a small effect size). This seeming disparity is due to the high class imbalance within safety reasons and white men's tendency to favor \emph{safe annotations}. And so for specific safety reasons they appear more agreeable. However, when these reasons are aggregated into an overall safety score, differences between which men and other groups reveal themselves.}
    \label{fig:Across-group-race-gender}
\end{figure*}

\begin{figure*}
    \centering
    \includegraphics[width=\textwidth]{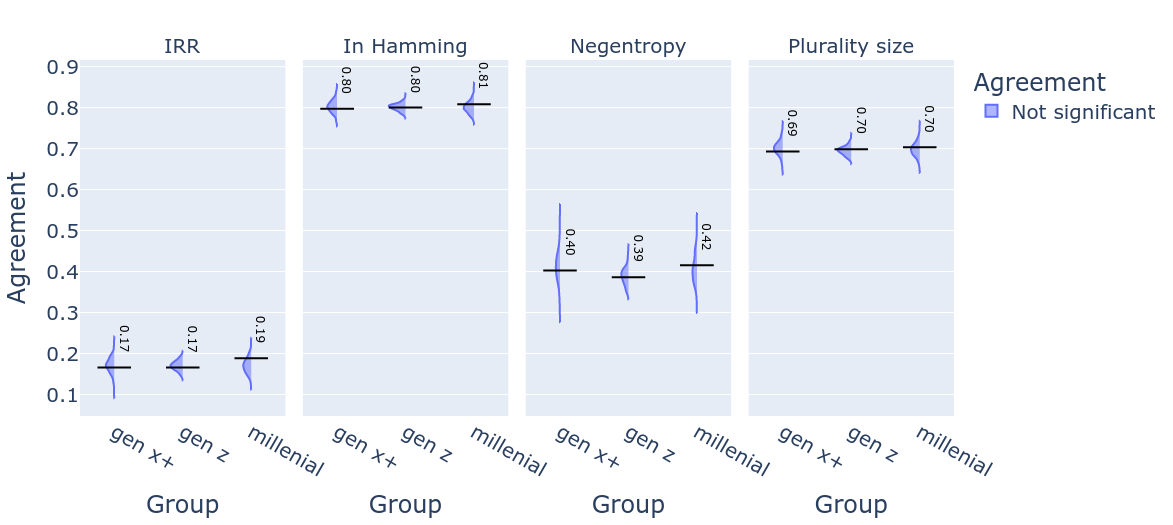}
    \caption{Within-group agreement metrics, by age. Histograms represent the distribution of agreement values under the null hypothesis. Black horizontal bars represent the observed values. None of these groups show significant amounts of difference in disagreement. } 
    \label{fig:within-group-age}
\end{figure*}

\begin{figure*}
    \centering
    \includegraphics[width=\textwidth]{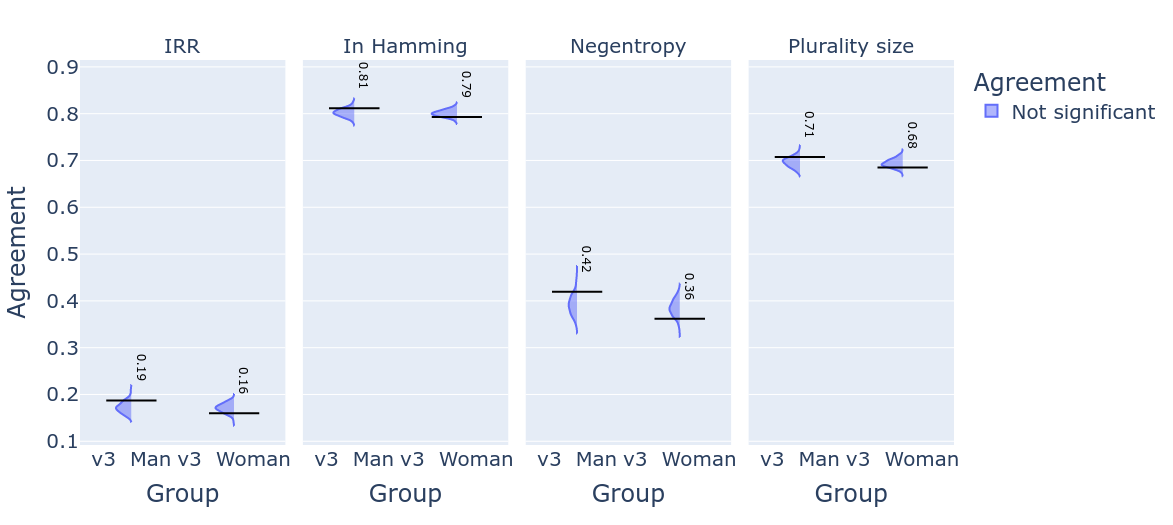}
    \caption{Within-group agreement metrics, by gender. Histograms represent the distribution of agreement values under the null hypothesis. Black horizontal bars represent the observed values.  None of these groups show significant amounts of difference in disagreement. }
    \label{fig:within-group-gender}
\end{figure*}

\begin{figure*}
    \centering
    \includegraphics[width=\textwidth]{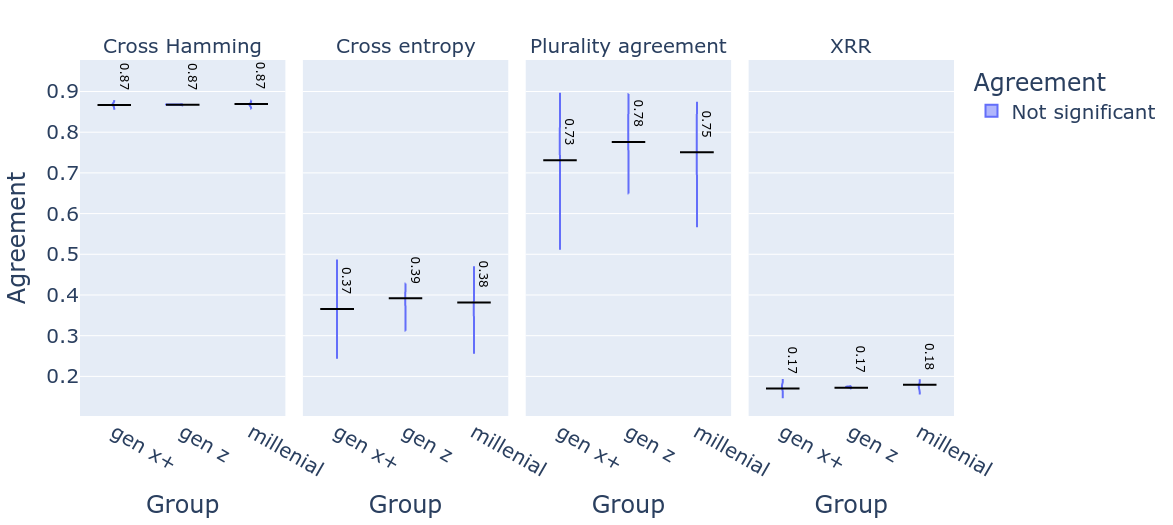}
    \caption{Across-group agreement metrics, by age. Histograms represent the distribution of agreement values under the null hypothesis. Black horizontal bars represent the observed values.}
    \label{fig:Across-group-age}
\end{figure*}

\begin{figure*}
    \centering
    \includegraphics[width=\textwidth]{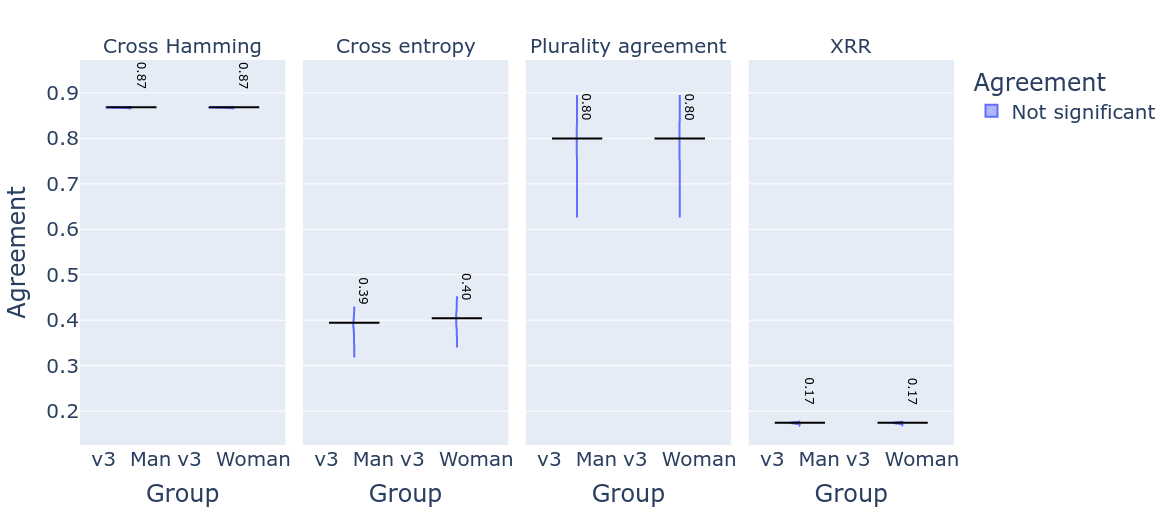}
    \caption{Across-group agreement metrics, by gender. Histograms represent the distribution of agreement values under the null hypothesis. Black horizontal bars represent the observed values.}
    \label{fig:Across-group-gender}
\end{figure*}

Figures \ref{fig:within-group-race}--\ref{fig:Across-group-gender} report, for each metric and demographic group, the score of the metric as a horizontal black line and, subimposed beneath each horizontal line, a histogram of the metric scores under the permutation sampling determined by our null hypothesis. Result are significant when the horizontal is at the extreme end of the  histograms. Histograms are also color-coded by the significance of the results they support: red histograms indicate that the result is significant at the $p=0.05$ level, but only before adjusting for the false positive rate (FPR); green indicates significance at the $p=0.05$ level, even after FPR adjustment. Given the exploratory nature of the work, both kinds of significance are meaningful and merit attention. But we can feel more confident that the FPR adjusted results are likely more robust and repeatable.

\begin{table*}
\scriptsize
\centering
\begin{tabular}{llllllllll}
\toprule
 &      &              &     &                   &                  &         Cross  &       Plurality &   Plurality  &                  \\
 &       Dimension &               Group &                  IRR &                  XRR &                        Negentropy&         Negentropy &        size &   agreement &                  GAI \\
\midrule
0  &           [age] &              gen x+ &               $\downarrow$0.166 &               $\downarrow$0.171 &               $\downarrow$0.402 &               $\downarrow$0.365 &               $\downarrow$0.693 &                $\downarrow$0.731 &               $\downarrow$0.975 \\
1  &           [age] &               gen z &               $\downarrow$0.166 &               $\downarrow$0.172 &               $\downarrow$0.386 &                 $\uparrow$0.392 &                 $\uparrow$0.698 &                $\downarrow$0.776 &               $\downarrow$0.966 \\
2  &           [age] &           millenial &                 $\uparrow$0.189 &                 $\uparrow$0.179 &                 $\uparrow$0.415 &                 $\uparrow$0.381 &                 $\uparrow$0.703 &                $\downarrow$0.751 &                 $\uparrow$1.052 \\
\midrule
3  &        [gender] &               Man &                 $\uparrow$0.187 &                 $\uparrow$0.175 &                 $\uparrow$0.419 &                 $\uparrow$0.394 &                 $\uparrow$0.707 &                  $\uparrow$0.800 &                 $\uparrow$1.071 \\
4  &        [gender] &             Woman &               $\downarrow$0.160 &                 $\uparrow$0.175 &               $\downarrow$0.362 &                 $\uparrow$0.404 &               $\downarrow$0.685 &                  $\uparrow$0.800 &               $\downarrow$0.916 \\
\midrule
5 &          [race] &               Asian &               $\downarrow$0.145 &               $\downarrow$0.166 &               $\downarrow$0.368 &               $\downarrow$0.323 &               $\downarrow$0.675 &                $\downarrow$0.740 &               $\downarrow$0.872 \\
6 &          [race] &               Black &                 $\uparrow$0.193 &                 $\uparrow$0.181 &               $\downarrow$0.411 &               $\downarrow$0.361 &                 $\uparrow$0.705 &                  $\uparrow$0.796 &                 $\uparrow$1.063 \\
7 &          [race] &              Latine &    \textbf{$\uparrow$0.215*} &    \textbf{$\uparrow$0.189*} &                 $\uparrow$0.467 &                 $\uparrow$0.412 &                 $\uparrow$0.716 &                $\downarrow$0.747 &    \textbf{$\uparrow$1.139*} \\
8 &          [race] &         Multiracial &               $\downarrow$0.153 &               $\downarrow$0.168 &  \textbf{$\downarrow$0.355*} &  \textbf{$\downarrow$0.250*} &  \textbf{$\downarrow$0.661*} &                $\downarrow$0.592 &               $\downarrow$0.916 \\
9 &          [race] &               White &               $\downarrow$0.145 &  \textbf{$\downarrow$0.159*} &    \textbf{$\uparrow$0.498*} &    \textbf{$\uparrow$0.417*} &    \textbf{$\uparrow$0.744*} &  \textbf{$\downarrow$0.552**} &               $\downarrow$0.908 \\
\midrule
10  &  [race, gender] &          Asian, Man &                 $\uparrow$0.193 &                 $\uparrow$0.188 &                 $\uparrow$0.495 &                 $\uparrow$0.417 &                 $\uparrow$0.733 &                  $\uparrow$0.722 &                 $\uparrow$1.024 \\
11  &  [race, gender] &        Asian, Woman &  \textbf{$\downarrow$0.073*} &  \textbf{$\downarrow$0.134*} &  \textbf{$\downarrow$0.332*} &  \textbf{$\downarrow$0.193*} &  \textbf{$\downarrow$0.633*} &                $\downarrow$0.543 &  \textbf{$\downarrow$0.540*} \\
12  &  [race, gender] &          Black, Man &               $\downarrow$0.139 &               $\downarrow$0.167 &               $\downarrow$0.502 &               $\downarrow$0.371 &               $\downarrow$0.710 &                $\downarrow$0.604 &               $\downarrow$0.831 \\
13  &  [race, gender] &        Black, Woman &    \textbf{$\uparrow$0.213*} &                 $\uparrow$0.188 &                 $\uparrow$0.441 &               $\downarrow$0.349 &                 $\uparrow$0.718 &                  $\uparrow$0.749 &    \textbf{$\uparrow$1.130*} \\
14  &  [race, gender] &         Latine, Man &                 $\uparrow$0.195 &                 $\uparrow$0.183 &                 $\uparrow$0.491 &                 $\uparrow$0.383 &                 $\uparrow$0.716 &                  $\uparrow$0.687 &                 $\uparrow$1.069 \\
15 &  [race, gender] &       Latine, Woman &    \textbf{$\uparrow$0.238*} &   \textbf{$\uparrow$0.199**} &                 $\uparrow$0.530 &                 $\uparrow$0.437 &                 $\uparrow$0.745 &                  $\uparrow$0.704 &    \textbf{$\uparrow$1.196*} \\
16 &  [race, gender] &    Multiracial, Man &                 $\uparrow$0.190 &                 $\uparrow$0.182 &               $\downarrow$0.432 &               $\downarrow$0.273 &               $\downarrow$0.688 &                $\downarrow$0.562 &                 $\uparrow$1.043 \\
17 &  [race, gender] &  Multiracial, Woman &               $\downarrow$0.041 &               $\downarrow$0.131 &  \textbf{$\downarrow$0.470*} &               $\downarrow$0.184 &               $\downarrow$0.674 &                $\downarrow$0.438 &               $\downarrow$0.312 \\
18 &  [race, gender] &          White, Man &    \textbf{$\uparrow$0.218*} &               $\downarrow$0.173 &   \textbf{$\uparrow$0.724**} &   \textbf{$\uparrow$0.505**} &   \textbf{$\uparrow$0.835**} &   \textbf{$\downarrow$0.446*} &   \textbf{$\uparrow$1.262**} \\
19 &  [race, gender] &        White, Woman &  \textbf{$\downarrow$0.114*} &  \textbf{$\downarrow$0.152*} &                 $\uparrow$0.454 &                 $\uparrow$0.381 &               $\downarrow$0.702 &                $\downarrow$0.663 &  \textbf{$\downarrow$0.752*} \\
\bottomrule
\end{tabular}

\caption{Results for in-group and cross-group cohesion, and GAI for demographic and intersectional groups within \textbf{DICES-350}. Significant results are in \textbf{bold}. A single asterisk (*) means the result is significant at the $p = 0.05$ level. A double asterisk (**) means the results are significant after Benjamini-Hochberg correction. A $\downarrow$ means that the result is less than expected under the null hypothesis. A $\uparrow$ means the result is greater. We report GAI based on $C_X=$ XRR and $C_I=$ IRR. The DSI results are based on variable that minimized each dimension, and they are as follows. Age: 1.052 (millennial), gender: 1.071 (men), race/ethnicity: 1.139 (Latine raters), (gender, race/ethnicity): 1.262 (White men).}
\label{tab:dices-results-all}
\end{table*}
\begin{table*}
\scriptsize
\centering
\begin{tabular}{llllllllll}

\toprule

 &&&&&&Cross  & Plurality & Plurality & \\
 &Dimension & Group &  IRR &  XRR & Negentropy & Negentropy & size & agreement & GAI \\
\midrule
0 & [age]&  (18,30)
&\textbf{$\uparrow$0.115**}
&$\uparrow$0.107
&\textbf{$\uparrow$0.631**}
&$\downarrow$0.297
&$\downarrow$0.405
&\textbf{$\downarrow$0.689**}
&\textbf{$\uparrow$1.068**}
\\
1 & [age]&  (30,50)
&\textbf{$\downarrow$0.089**}
&$\downarrow$0.104
&\textbf{$\downarrow$0.571**}
&$\uparrow$0.340
&\textbf{$\downarrow$0.377*}
&\textbf{$\uparrow$0.720**}
&\textbf{$\downarrow$0.850**}
\\
2 & [age]&  50+
&$\uparrow$0.110
&$\uparrow$0.111
&$\downarrow$0.480
&$\uparrow$0.389
&$\uparrow$0.356
&$\uparrow$0.754
&$\uparrow$0.999
\\\midrule
3 & [gender]&  Woman
&$\uparrow$0.110
&$\uparrow$0.108
&\textbf{$\uparrow$0.634**}
&\textbf{$\downarrow$0.267**}
&$\uparrow$0.424
&\textbf{$\downarrow$0.692**}
&$\uparrow$1.024\\
4 & [gender]&  Man
&$\downarrow$0.105
&$\uparrow$0.107
&\textbf{$\downarrow$0.612**}
&\textbf{$\uparrow$0.307**}
&$\uparrow$0.423
&\textbf{$\uparrow$0.702**}
&$\downarrow$0.976
\\
5 & [gender]&  Other
&$\uparrow$0.209
&$\downarrow$0.096
&$\downarrow$0.030
&$\downarrow$0.605
&$\uparrow$0.192
&$\uparrow$0.978
&\textbf{$\uparrow$2.172*}
\\\midrule
6 & [region]&  Arab Culture
&\textbf{$\uparrow$0.133**}
&$\uparrow$0.113
&\textbf{$\uparrow$0.452**}
&$\downarrow$0.413
&$\downarrow$0.272
&\textbf{$\downarrow$0.759**}
&\textbf{$\uparrow$1.174*}
\\
7 & [region]&  Indian Cultural Sphere
&$\downarrow$0.103
&\textbf{$\downarrow$0.099*}
&\textbf{$\uparrow$0.457**}
&$\downarrow$0.418
&$\downarrow$0.280
&\textbf{$\downarrow$0.760**}
&$\uparrow$1.043
\\
8 & [region]&  Latin America
&\textbf{$\uparrow$0.129**}
&$\uparrow$0.112
&\textbf{$\uparrow$0.449**}
&\textbf{$\downarrow$0.400*}
&$\uparrow$0.317
&\textbf{$\downarrow$0.764**}
&\textbf{$\uparrow$1.152*}
\\
9 & [region]&  North America
&\textbf{$\uparrow$0.143**}
&$\uparrow$0.110
&\textbf{$\uparrow$0.443**}
&\textbf{$\downarrow$0.393**}
&$\uparrow$0.316
&$\downarrow$0.772
&\textbf{$\uparrow$1.307**}
\\
10 & [region]&  Oceania
&$\uparrow$0.118
&$\downarrow$0.103
&\textbf{$\downarrow$0.372**}
&$\downarrow$0.411
&$\uparrow$0.303
&\textbf{$\uparrow$0.797**}
&\textbf{$\uparrow$1.145*}
\\
11 & [region]&  Sinosphere
&\textbf{$\downarrow$0.087*}
&\textbf{$\downarrow$0.087**}
&$\downarrow$0.405
&\textbf{$\downarrow$0.381**}
&\textbf{$\downarrow$0.223**}
&$\uparrow$0.788
&$\downarrow$1.002
\\
12 & [region]&  Sub Saharan Africa
&\textbf{$\uparrow$0.142**}
&$\downarrow$0.104
&$\uparrow$0.418
&\textbf{$\downarrow$0.385**}
&\textbf{$\downarrow$0.262*}
&$\downarrow$0.777
&\textbf{$\uparrow$1.361**}
\\
13 & [region]&  Western Europe
&\textbf{$\uparrow$0.135**}
&$\uparrow$0.111
&\textbf{$\uparrow$0.448**}
&\textbf{$\downarrow$0.383**}
&\textbf{$\uparrow$0.356**}
&$\downarrow$0.768
&\textbf{$\uparrow$1.222**}
\\
\bottomrule
 \end{tabular}
\caption{Results for in-group and cross-group cohesion, and GAI for demographic groups of \textbf{D3} raters. Significant results are in \textbf{bold}: * for significance at $p < 0.05$, ** for significance after Benjamini-Hochberg correction.
A single asterisk (*) means significant at the $p = 0.05$ level. A double asterisk (**) means the results are significant after Benjamini-Hochberg correction. 
A $\downarrow$ (or $\uparrow$) means that the result is less (or greater) than expected under the null hypothesis. 
GAI results based on $C_X=$ XRR and $C_I=$ IRR.}
\label{tab:d3-results-all}
\end{table*}

\begin{table*}
\scriptsize
\centering
\begin{tabular}{llllllllll}

\toprule

 &&&&&&Cross  & Plurality & Plurality & \\
 &Dimension & Group &  IRR &  XRR & Negentropy & Negentropy & size & agreement & GAI \\
\midrule
0 & [region, age]& AC., (18,30)
&$\uparrow$0.119
&$\uparrow$0.111
&$\uparrow$0.268
&$\uparrow$0.477
&\textbf{$\downarrow$0.207*}
&$\downarrow$0.836
&$\uparrow$1.070
\\
1 & [region, age]& AC., (30,50)
&$\uparrow$0.116
&$\uparrow$0.112
&$\downarrow$0.184
&$\downarrow$0.481
&$\downarrow$0.226
&$\downarrow$0.886
&$\uparrow$1.040
\\
2 & [region, age]& AC., 50+
&\textbf{$\uparrow$0.190*}
&\textbf{$\uparrow$0.179**}
&$\downarrow$0.080
&\textbf{$\uparrow$0.610**}
&$\uparrow$0.228
&$\uparrow$0.947
&$\uparrow$1.060
\\
3 & [region, gender]& AC., Man
&$\uparrow$0.129
&$\uparrow$0.109
&$\downarrow$0.284
&\textbf{$\uparrow$0.489**}
&$\downarrow$0.227
&$\downarrow$0.828
&$\uparrow$1.185
\\
4 & [region, gender]& AC., Woman
&$\uparrow$0.125
&$\uparrow$0.117
&$\downarrow$0.198
&$\uparrow$0.488
&$\downarrow$0.202
&$\uparrow$0.875
&$\uparrow$1.064
\\\midrule
5 & [region, age]& ICS., (18,30)
&\textbf{$\downarrow$0.063**}
&$\downarrow$0.100
&$\uparrow$0.246
&$\uparrow$0.485
&$\downarrow$0.223
&$\downarrow$0.849
&\textbf{$\downarrow$0.634*}
\\
6 & [region, age]& ICS., (30,50)
&\textbf{$\downarrow$0.060*}
&$\downarrow$0.100
&$\uparrow$0.215
&$\downarrow$0.482
&$\uparrow$0.236
&$\downarrow$0.868
&\textbf{$\downarrow$0.601*}
\\
7 & [region, age]& ICS., 50+
&$\downarrow$0.063
&$\downarrow$0.103
&$\downarrow$0.121
&$\downarrow$0.513
&$\uparrow$0.246
&$\uparrow$0.922
&$\downarrow$0.614
\\
8 & [region, gender]& ICS., Man
&$\downarrow$0.093
&$\downarrow$0.098
&$\downarrow$0.284
&$\downarrow$0.455
&$\downarrow$0.241
&$\downarrow$0.831
&$\downarrow$0.953
\\
9 & [region, gender]& ICS., Woman
&\textbf{$\downarrow$0.070*}
&$\downarrow$0.106
&$\downarrow$0.233
&$\uparrow$0.475
&\textbf{$\downarrow$0.197**}
&$\uparrow$0.860
&\textbf{$\downarrow$0.655*}
\\\midrule
10 & [region, age]& LA., (18,30)
&\textbf{$\uparrow$0.143**}
&$\uparrow$0.118
&$\downarrow$0.278
&$\uparrow$0.475
&$\downarrow$0.248
&$\uparrow$0.837
&\textbf{$\uparrow$1.216*}
\\
11 & [region, age]& LA., (30,50)
&$\downarrow$0.069
&\textbf{$\downarrow$0.092*}
&\textbf{$\uparrow$0.227**}
&$\uparrow$0.514
&$\downarrow$0.209
&\textbf{$\downarrow$0.864*}
&$\downarrow$0.747
\\
12 & [region, age]& LA., 50+
&$\uparrow$0.158
&$\uparrow$0.136
&$\uparrow$0.096
&$\uparrow$0.583
&$\uparrow$0.235
&$\downarrow$0.933
&$\uparrow$1.157
\\
13 & [region, gender]& LA., Man
&$\uparrow$0.118
&$\downarrow$0.108
&$\downarrow$0.259
&$\uparrow$0.477
&$\downarrow$0.228
&$\downarrow$0.842
&$\uparrow$1.096
\\
14 & [region, gender]& LA., Woman
&\textbf{$\uparrow$0.143**}
&$\uparrow$0.111
&$\downarrow$0.251
&$\uparrow$0.473
&$\downarrow$0.241
&$\uparrow$0.849
&\textbf{$\uparrow$1.290*}
\\\midrule
15 & [region, age]& NA., (18,30)
&\textbf{$\uparrow$0.150**}
&\textbf{$\uparrow$0.124**}
&$\uparrow$0.272
&$\uparrow$0.472
&$\uparrow$0.250
&\textbf{$\downarrow$0.829**}
&$\uparrow$1.215
\\
16 & [region, age]& NA., (30,50)
&$\uparrow$0.105
&$\downarrow$0.102
&$\downarrow$0.173
&$\downarrow$0.471
&$\uparrow$0.249
&\textbf{$\uparrow$0.898*}
&$\uparrow$1.024
\\
17 & [region, age]& NA., 50+
&$\downarrow$0.099
&$\downarrow$0.098
&$\uparrow$0.139
&$\downarrow$0.519
&$\downarrow$0.210
&$\downarrow$0.911
&$\uparrow$1.016
\\
18 & [region, gender]& NA., Man
&$\uparrow$0.113
&$\uparrow$0.112
&\textbf{$\downarrow$0.188**}
&\textbf{$\downarrow$0.454*}
&\textbf{$\uparrow$0.278*}
&\textbf{$\uparrow$0.885**}
&$\uparrow$1.005
\\
19 & [region, gender]& NA., Woman
&\textbf{$\uparrow$0.153**}
&$\uparrow$0.116
&$\uparrow$0.299
&$\downarrow$0.449
&$\downarrow$0.239
&$\downarrow$0.825
&\textbf{$\uparrow$1.314**}
\\\midrule
20 & [region, age]& Oc., (18,30)
&$\uparrow$0.113
&$\uparrow$0.121
&$\downarrow$0.155
&$\uparrow$0.510
&$\uparrow$0.230
&$\uparrow$0.900
&$\uparrow$0.932
\\
21 & [region, age]& Oc., (30,50)
&$\uparrow$0.112
&\textbf{$\downarrow$0.089**}
&\textbf{$\downarrow$0.173**}
&\textbf{$\downarrow$0.455*}
&$\downarrow$0.218
&\textbf{$\uparrow$0.900**}
&\textbf{$\uparrow$1.255*}
\\
22 & [region, age]& Oc., 50+
&$\downarrow$0.081
&$\uparrow$0.115
&$\downarrow$0.140
&\textbf{$\downarrow$0.481*}
&\textbf{$\uparrow$0.286**}
&$\uparrow$0.914
&$\downarrow$0.699
\\
23 & [region, gender]& Oc., Man
&$\downarrow$0.090
&\textbf{$\downarrow$0.091*}
&\textbf{$\downarrow$0.170**}
&\textbf{$\downarrow$0.448**}
&$\downarrow$0.219
&\textbf{$\uparrow$0.899**}
&$\uparrow$0.988
\\
24 & [region, gender]& Oc., Woman
&\textbf{$\uparrow$0.133*}
&$\uparrow$0.110
&\textbf{$\downarrow$0.252**}
&$\uparrow$0.464
&$\uparrow$0.266
&\textbf{$\uparrow$0.853**}
&\textbf{$\uparrow$1.208*}
\\\midrule
25 & [region, age]& Si., (18,30)
&$\uparrow$0.112
&$\downarrow$0.108
&$\downarrow$0.190
&\textbf{$\downarrow$0.456**}
&$\downarrow$0.217
&$\uparrow$0.883
&$\uparrow$1.029
\\
26 & [region, age]& Si., (30,50)
&\textbf{$\downarrow$0.033**}
&\textbf{$\downarrow$0.082**}
&\textbf{$\downarrow$0.209*}
&\textbf{$\downarrow$0.423**}
&\textbf{$\downarrow$0.175**}
&$\uparrow$0.873
&\textbf{$\downarrow$0.405**}
\\
27 & [region, age]& Si., 50+
&$\uparrow$0.137
&\textbf{$\downarrow$0.061**}
&\textbf{$\downarrow$0.071**}
&\textbf{$\downarrow$0.478**}
&\textbf{$\downarrow$0.152**}
&\textbf{$\uparrow$0.954**}
&\textbf{$\uparrow$2.225**}
\\
28 & [region, gender]& Si., Man
&$\downarrow$0.093
&\textbf{$\downarrow$0.091**}
&$\downarrow$0.260
&\textbf{$\downarrow$0.426**}
&\textbf{$\downarrow$0.190**}
&$\uparrow$0.843
&$\uparrow$1.022
\\
29 & [region, gender]& Si., Woman
&$\downarrow$0.100
&\textbf{$\downarrow$0.081**}
&\textbf{$\downarrow$0.196**}
&\textbf{$\downarrow$0.413**}
&\textbf{$\downarrow$0.168**}
&\textbf{$\uparrow$0.883**}
&\textbf{$\uparrow$1.237*}
\\\midrule
30 & [region, age]& SSA., (18,30)
&\textbf{$\uparrow$0.146**}
&$\downarrow$0.107
&$\downarrow$0.280
&$\uparrow$0.462
&\textbf{$\downarrow$0.222*}
&$\uparrow$0.834
&\textbf{$\uparrow$1.365**}
\\
31 & [region, age]& SSA., (30,50)
&$\uparrow$0.135
&$\uparrow$0.119
&$\downarrow$0.160
&$\downarrow$0.485
&$\downarrow$0.218
&$\uparrow$0.900
&$\uparrow$1.137
\\
32 & [region, age]& SSA., 50+
&$\uparrow$0.163
&$\uparrow$0.125
&$\uparrow$0.079
&$\downarrow$0.592
&$\uparrow$0.208
&$\downarrow$0.950
&$\uparrow$1.299
\\
33 & [region, gender]& SSA., Man
&\textbf{$\uparrow$0.132*}
&\textbf{$\uparrow$0.119*}
&$\downarrow$0.286
&\textbf{$\downarrow$0.435*}
&$\uparrow$0.268
&$\downarrow$0.829
&$\uparrow$1.104
\\
34 & [region, gender]& SSA., Woman
&$\uparrow$0.119
&$\uparrow$0.109
&$\downarrow$0.213
&$\downarrow$0.470
&$\downarrow$0.233
&$\uparrow$0.870
&$\uparrow$1.093
\\\midrule
35 & [region, age]& WE., (18,30)
&\textbf{$\uparrow$0.177**}
&\textbf{$\uparrow$0.126**}
&$\downarrow$0.246
&$\uparrow$0.469
&\textbf{$\uparrow$0.285*}
&$\downarrow$0.849
&\textbf{$\uparrow$1.402**}
\\
36 & [region, age]& WE., (30,50)
&$\downarrow$0.085
&\textbf{$\downarrow$0.093*}
&$\downarrow$0.173
&$\downarrow$0.487
&$\downarrow$0.205
&$\uparrow$0.896
&$\downarrow$0.923
\\
37 & [region, age]& WE., 50+
&$\uparrow$0.117
&$\downarrow$0.104
&$\uparrow$0.152
&$\uparrow$0.545
&$\uparrow$0.220
&$\downarrow$0.905
&$\uparrow$1.120
\\
38 & [region, gender]& WE., Man
&$\uparrow$0.116
&$\downarrow$0.106
&\textbf{$\downarrow$0.214**}
&\textbf{$\downarrow$0.443**}
&$\uparrow$0.257
&\textbf{$\uparrow$0.874**}
&$\uparrow$1.096
\\
39 & [region, gender]& WE., Woman
&\textbf{$\uparrow$0.151**}
&$\uparrow$0.118
&$\uparrow$0.292
&$\downarrow$0.452
&$\downarrow$0.243
&\textbf{$\downarrow$0.825*}
&\textbf{$\uparrow$1.284*}
\\
\bottomrule
 \end{tabular}
\caption{Results for in-group and cross-group cohesion, and GAI for intersectional demographic groups within \textbf{D3}. Significant results are in \textbf{bold}: * for significance at $p < 0.05$, ** for significance after Benjamini-Hochberg correction.
A single asterisk (*) means significant at the $p = 0.05$ level. A double asterisk (**) means the results are significant after Benjamini-Hochberg correction. 
A $\downarrow$ (or $\uparrow$) means that the result is less (or greater) than expected under the null hypothesis. 
GAI results based on $C_X=$ XRR and $C_I=$ IRR.}
\label{tab:d3-interactions-results}
\end{table*}

\end{document}